\documentclass[10pt,twocolumn,letterpaper]{article}

\usepackage{cvpr}              
\usepackage{makecell}
\usepackage{amsmath}  
\usepackage{amssymb}  
\usepackage{booktabs}
\usepackage{xspace}
\usepackage{multirow}

\newcommand{\algname} {Med3D-R1\xspace}

\definecolor{cvprblue}{rgb}{0.21,0.49,0.74}
\usepackage[pagebackref,breaklinks,colorlinks,allcolors=cvprblue]{hyperref}

\def\paperID{} 
\def\confName{CVPR}
\def\confYear{2026}

\title{\algname: Incentivizing Clinical Reasoning in 3D Medical Vision-Language Models for Abnormality Diagnosis}

\author{
    Haoran Lai$^{1,2,}$\qquad 
    Zihang Jiang$^{2,}$\thanks{Corresponding author} \qquad
    Kun Zhang$^{2}$\qquad
    Qingsong Yao$^{3}$\qquad 
    Rongsheng Wang$^{2}$ \\
    Zhiyang He$^{4}$ \qquad 
    Xiaodong Tao$^{4}$ \qquad 
    Wei Wei$^{5}$ \qquad 
    S. Kevin Zhou$^{1,2,3,}$\footnotemark[1] \\
$^{1}$ School of Biomedical Engineering, Division of Life Sciences and Medicine, \\ 
University of Science and Technology of China, Hefei, Anhui, 230026, P.R.China \\
$^{2}$ Suzhou Institute for Advanced Research, University of Science and  Technology of  China, \\ Suzhou, Jiangsu, 215123, P.R.China \\
$^{3}$  Stanford University, Palo Alto, California, 94305, United State \\
$^{4}$ Medical Business Department, iFlytek Co.Ltd, Hefei 230088, China \\
$^{5}$ The First Affiliated Hospital of USTC, Division of Life Sciences and Medicine \\ University of Science and Technology of China, Hefei, Anhui, 230001, China \\
{\tt\small \{haoranlai, rongsheng\_wang\}@mail.ustc.edu.cn} \\
{\tt\small \{zyh, xdtao\}@iflytek.com, jzh0103@ustc.edu.cn, s.kevin.zhou@gmail.com} }

\begin{document}

\maketitle
\begin{abstract}
Developing 3D vision-language models with robust clinical reasoning remains a challenge due to the inherent complexity of volumetric medical imaging, the tendency of models to overfit superficial report patterns, and the lack of interpretability-aware reward designs. In this paper, we propose \algname, a reinforcement learning framework with a two-stage training process: Supervised Fine-Tuning (SFT) and Reinforcement Learning (RL). During SFT stage, we introduce a residual alignment mechanism to bridge the gap between high-dimensional 3D features and textual embeddings, and an abnormality re-weighting strategy to emphasize clinically informative tokens and reduce structural bias in reports. In RL stage, we redesign the consistency reward to explicitly promote coherent, step-by-step diagnostic reasoning. We evaluate our method on medical multiple-choice visual question answering using two 3D diagnostic benchmarks, CT-RATE and RAD-ChestCT, where our model attains state-of-the-art accuracies of 41.92\% on CT-RATE and 44.99\% on RAD-ChestCT. These results indicate improved abnormality diagnosis and clinical reasoning and outperform prior methods on both benchmarks. Overall, our approach holds promise for enhancing real-world diagnostic workflows by enabling more reliable and transparent 3D medical vision-language systems.
\end{abstract}    
\section{Introduction}
\label{sec:intro}

With the growing availability of volumetric medical imaging data, 3D diagnosis based on Computed Tomography (CT) has become a crucial task in clinical decision-making~\cite{ginat2014advances,oikonomou2018non,bera2019artificial}. Recently, Large Language Models (LLMs) have been increasingly adopted to enhance diagnostic reasoning by unifying vision and language modalities, heralding a new paradigm for medical artificial intelligence~\cite{pan2025medvlm,lai2025med}. Under this paradigm~\cite{ouyang2022training}, a mainstream two-stage training strategy is typically employed: first, \emph{Supervised Fine-Tuning} (SFT) is used to align visual and textual features while concurrently endowing the model with specialized medical knowledge needed for key tasks; second, \emph{Reinforcement Learning} (RL) further refines the model’s outputs to meet specific clinical objectives and improve real-world applicability.

\begin{figure}[t]
  \centering
    \includegraphics[width=\linewidth]{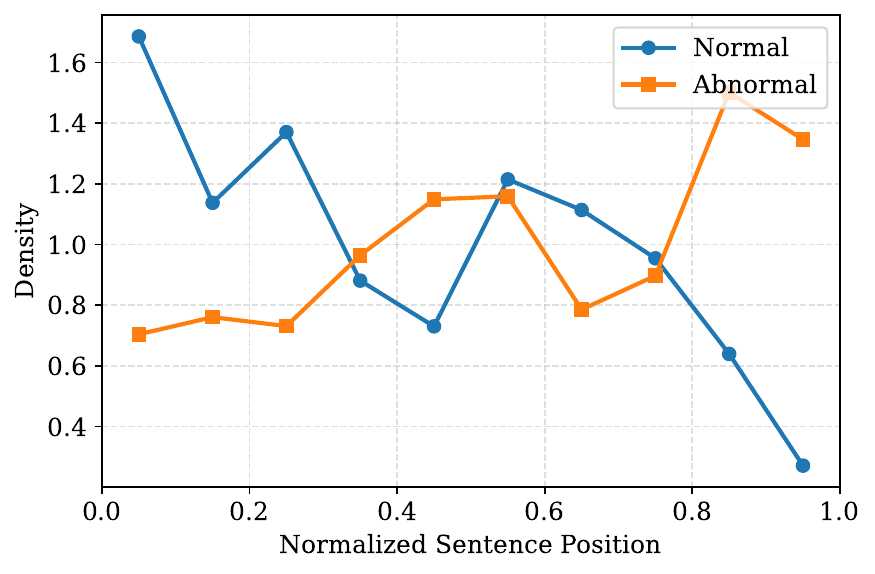}
    \caption{
    Density distribution of normal and abnormal sentences over normalized report positions.
    Normal sentences predominantly appear at the beginning of reports, whereas abnormal sentences are increasingly concentrated toward later positions, revealing a consistent positional bias in radiology report writing.
    }

    \label{fig:data_distribution}
\end{figure}

However, despite their promising potential, existing methods under this paradigm still suffer from several key limitations when applied to 3D medical imaging. First, in contrast to 2D images, 3D volumetric data presents substantially greater complexity in spatial structure, scale variance, and semantic granularity~\cite{lin2024ct,shui2025large}. This complexity significantly complicates the extraction of diagnostically salient features and their alignment with corresponding textual representations. The challenge is further compounded by the nature of clinical reports, which predominantly consist of high-level abstractions and often lack detailed spatial annotations. Such semantic abstraction in textual descriptions exacerbates the misalignment between high-dimensional visual inputs and abstract language outputs, ultimately leading to unstable optimization and suboptimal performance in diagnostic reasoning.

Second, we find that SFT strategies often \emph{overfit to superficial patterns}, with models relying heavily on frequently repeated and formulaic language structures in the training set rather than developing genuine reasoning abilities. While the overall class balance between normal and abnormal cases is relatively moderate, a more critical issue lies in the structural composition of medical reports. As shown in Figure~\ref{fig:data_distribution}, normal findings are predominantly concentrated in the early sections of the report, whereas abnormal findings are more frequently positioned toward the latter parts. This asymmetric distribution reveals a fixed ``normal-first'' narrative pattern, where templated normal descriptions tend to precede more variable abnormal observations. Such reporting conventions may implicitly bias the model toward defaulting to normal predictions early on, thereby impairing its ability to identify and reason about clinically meaningful abnormalities that break this learned template.

Furthermore, the current RL strategies largely overlook the reasoning process underlying clinical diagnosis. Existing reward functions~\cite{pan2025medvlm,lai2025med} focus mostly on the final output correctness without modeling the intermediate decision-making steps that mirror how radiologists analyze and deduce from 3D volumes. This lack of interpretability-aware reward design leads to suboptimal learning of reasoning chains and weakens the overall diagnostic consistency and reliability.

To address these limitations, we introduce \algname, a unified framework for 3D medical vision-language modeling that explicitly enhances clinical reasoning capabilities. In the SFT stage, we propose two key modules: the Residual Alignment Mechanism (RAM), which bridges the representational gap between high-dimensional 3D features and the textual embedding space; and the Abnormality Re-Weighting (ARW) strategy, which prioritizes clinically significant tokens to counteract biases introduced by the structured narrative style of radiology reports. To further strengthen the model’s reasoning ability, we introduce a RL stage that directly optimizes diagnostic correctness and
coherence. Unlike prior methods~\cite{lai2025med,pan2025medvlm} our RL design includes a novel consistency reward that aligns reasoning chains with clinical logic, enabling more trustworthy
decision-making. Together, these improvements form a cohesive framework that significantly enhances both the interpretability and diagnostic performance of 3D vision-language models.

To validate our approach, we conduct extensive experiments on two diagnostic benchmarks—CT-RATE~\cite{hamamci2024developing} and RAD-ChestCT~\cite{draelos2021machine}—under the Medical Multiple-choice Visual Question Answering (MMVQA) setting. Our model attains accuracies of 41.92\% on CT-RATE and 44.99\% on RAD-ChestCT, achieving state-of-the-art results on both benchmarks. These results indicate strong capability in abnormality diagnosis and clinical reasoning, and further highlight the method’s potential to support real-world diagnostic workflows by bridging the gap between raw volumetric data and interpretable medical decision-making.

Our main contributions are summarized as follows:
\begin{itemize}

    \item To the best of our knowledge, we are the first to show that the ordering of normal and abnormal statements in structured reports induces a positional bias harmful to autoregressive training, and we propose \emph{ARW} to alleviate it.

    \item We apply RL to 3D medical VLMs for the first time, introducing a \emph{consistency reward} to promote clinically grounded and coherent reasoning, ensuring more trustworthy inferences.
    
    \item Our approach attains 41.92\% and 44.99\% accuracy  on CT-RATE and RAD-ChestCT, respectively, establishing new state-of-the-art results under the MMVQA setting.

\end{itemize}

\section{Related Work}
\label{sec:relatedwork}

\subsection{Medical Vision-Language Models}

VLMs have demonstrated significant potential across a range of medical imaging tasks, including report generation, Visual Question Answering (VQA), and image-text retrieval. Early efforts such as CheXzero~\cite{tiu2022expert} and CARZero~\cite{lai2024carzero} employed contrastive learning to align radiological images with corresponding reports, primarily targeting 2D modalities like chest X-rays.  Building upon these foundations, recent efforts have advanced toward more comprehensive medical VLMs by integrating large language models with visual encoders. Representative models such as Med-Flamingo~\cite{moor2023med}, LLaVA-Med~\cite{li2024llava}, Med-PaLM~\cite{tu2024towards}, MedGemma~\cite{sellergren2025medgemma}, Lingshu~\cite{xu2025lingshu}, and HuatuoGPT-Vision~\cite{chen2024towards} extend general-purpose architectures to the medical domain via instruction tuning, enabling more expressive understanding, fine-grained reasoning, and multi-task generalization.

While these approaches have significantly advanced multimodal understanding in the medical domain, they predominantly focus on 2D imaging or tasks centered on surface-level vision-language alignment. Compared to 2D images, 3D medical imaging modalities such as CT present new challenges, including high dimensionality, spatial redundancy, and increased diagnostic complexity. In response, an increasing number of studies have begun to investigate how to effectively model and align 3D data within vision-language frameworks. Notable examples include RadFM~\cite{wu2025towards}, M3D~\cite{bai2024m3d}, E3D-GPT~\cite{lai2024e3d}, CT-CHAT~\cite{hamamci2024developing}, and Med3DVLM~\cite{xin2025med3dvlm}, which demonstrate the feasibility of extending vision-language modeling into the 3D space, achieving encouraging performance on downstream tasks.

Despite recent progress, current 3D medical vision-language models remain limited in their ability to perform clinical reasoning. Most existing approaches rely solely on SFT, which often leads to overfitting on surface-level patterns in the training data. This not only weakens the language understanding capabilities of the underlying LLMs but also fails to cultivate structured diagnostic reasoning. As a result, models tend to prioritize pattern memorization over genuine decision-making processes. In contrast, our work takes a significant step forward by incorporating RL to explicitly enhance reasoning ability, enabling more faithful, interpretable, and clinically grounded predictions in the 3D setting.

\begin{figure*}[htpb]
  \centering
  \includegraphics[width=\linewidth]{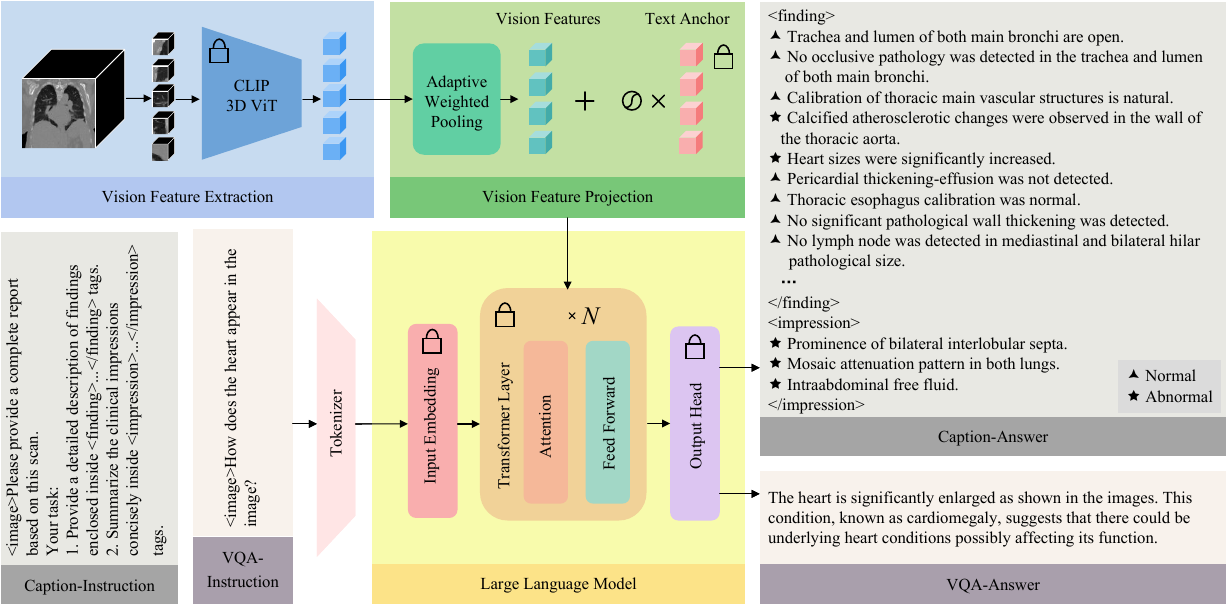}
  \caption{Overview of the SFT framework. A residual alignment mechanism refines the mapping of 3D image features from the ViT to a fixed text anchor, thereby enhancing cross-modal alignment. An abnormality re-weighting strategy further emphasizes clinically significant findings in the generation loss. During this stage, only the adaptive weighted pooling module is trainable, while all other components remain frozen.}
    \label{fig:sft_stage1}
\end{figure*}

\subsection{3D Medical Vision-Language Alignment}

Aligning 3D medical images with corresponding textual semantics remains a fundamental yet challenging problem in multimodal learning. Recent methods have explored various strategies to project volumetric features into the language space, typically relying on direct mappings from 3D visual embeddings to text representations. RadFM~\cite{wu2025towards} employs a Qformer-based architecture to aggregate 3D features before alignment. M3D~\cite{bai2024m3d} introduces spatial average pooling to compress volumetric features, while E3D-GPT~\cite{lai2024e3d} enhances this with adaptive weighted pooling for improved representation. Med3DVLM~\cite{xin2025med3dvlm} further applies an MLP-Mixer to integrate spatial and semantic information across slices. In addition to these direct alignment strategies, some recent works incorporate auxiliary mechanisms to ease the learning burden. For instance, 3DHRG~\cite{liu2024benchmarking} employs low-resolution volumes to guide high-resolution feature learning, and Med-2E3~\cite{shi2024med} uses 2D image-text alignment to inform 3D slice selection, offering a form of indirect supervision.

While effective to some extent, these approaches typically learn a direct projection from high-dimensional image features to full-text embeddings. Such mappings can be unstable or inefficient for 3D medical data, which is often sparse, structurally complex, and semantically diverse. To address this issue, we propose a residual alignment mechanism that introduces learnable text anchors as intermediate semantic targets. Rather than aligning directly to entire textual representations, the model learns to map 3D features onto residuals relative to these anchors, effectively transforming the alignment task into a localized refinement process. This design reduces the difficulty of optimization, enhances semantic interpretability, and facilitates more accurate vision-language alignment in the 3D setting.

\subsection{Medical Reasoning and Interpretability}

Reasoning plays a central role in medical decision-making, where models must not only provide correct answers but also generate clinically consistent justifications. While SFT has been the mainstream approach to aligning VLMs with medical tasks, it often encourages shortcut learning and fails to capture the underlying logic of clinical workflows. This issue is particularly severe in medical settings, where explainability and domain alignment are essential.

To overcome these limitations, recent studies have introduced  RL to improve reasoning generalization in 2D medical vision-language models. Works such as MedVLM-R1~\cite{pan2025medvlm} and Med-R1~\cite{lai2025med} demonstrate that reward-guided learning can enhance both interpretability and robustness across modalities and tasks. By leveraging Group Relative Policy Optimization (GRPO)~\cite{shao2024deepseekmath,shen2025vlmr1}, these models learn to produce structured rationales aligned with clinical guidelines, even in the absence of explicit chain-of-thought supervision. Rather than optimizing solely for answer accuracy, GRPO-based RL encourages exploration of diverse reasoning paths that are both correct and logically coherent.

However, these advances have so far been limited to 2D domains such as chest X-rays or dermatology images. To the best of our knowledge, our work is the first to extend reinforcement learning-based reasoning supervision to the 3D medical vision-language setting. Beyond the commonly used format and accuracy rewards, we introduce a novel consistency reward, which directly supervises the semantic alignment between the model’s intermediate reasoning and the reference radiology report. This design fosters the generation of faithful, structured, and clinically grounded explanations, bridging the gap between volumetric image interpretation and reliable diagnostic decision-making.
\section{Method}
\label{sec:method}

We propose \algname, a 3D medical VLM designed to improve both diagnostic accuracy and interpretability through three complementary modules. First, the Residual Alignment Mechanism (RAM) enhances cross-modal alignment by interpolating image features toward a fixed text anchor. Second, the Abnormality Re-weighting (ARW) strategy rebalances the training loss during SFT to emphasize clinically significant abnormal findings, as illustrated in Figure~\ref{fig:sft_stage1}. Finally, in the reinforcement learning (RL) stage, we introduce a consistency-guided reward to explicitly encourage accurate and clinically grounded diagnostic reasoning. We elaborate on each component below.

\subsection{Residual Alignment Mechanisms}

3D CT volumes carry rich anatomical and pathological information, leading to a high-dimensional and diverse visual feature space. In contrast, clinical reports are relatively sparse and exhibit lower variance, making it challenging to directly align these two modalities.

To bridge this gap, we first apply an adaptive weighted pooling module~\cite{lai2024e3d} to the features extracted by 3D vision encoder. This module learns to aggregate spatially distributed tokens into a more compact set of representations that capture the most diagnostically relevant cues. Building on this, we propose a RAM that reframes alignment as learning the \emph{difference} between image features and a fixed semantic anchor in text space, rather than performing a direct mapping. Inspired by residual learning~\cite{he2016deep}, where $h(x)=x+f(x)$ typically proves easier to optimize than $h(x)=f(x)$ alone, we focus on residual transformations to reduce the difficulty of bridging 3D image embeddings and lower-dimensional text embeddings.

Concretely, let \(E \in \mathbb{R}^{V \times D}\) be the word-embedding weight matrix obtained from a pretrained language model’s tokenizer.  Each row vector \(E_{j}\in\mathbb{R}^{D}\) is the \(D\)-dimensional
embedding of the \(j\)-th vocabulary token, and \(V\) is the vocabulary
size. We compute a single \textbf{text anchor}
\begin{equation}
t_{\text{anchor}}
    = \frac{1}{V}\sum_{j=1}^{V} E_{j}
    \;\in\; \mathbb{R}^{D},
\end{equation}
i.e., the mean vector of all vocabulary embeddings, and keep it fixed
throughout training.

For each projected image token \(x_i \in \mathbb{R}^{D}\), the RAM interpolates between \(x_i\) and
\(t_{\text{anchor}}\) via a token-specific gate:
\begin{equation}
z_i = \sigma_i \, t_{\text{anchor}}
      + (1-\sigma_i) \, x_i,
\end{equation}
\begin{equation}
\sigma_i
    = \operatorname{Sigmoid}\!\bigl(\operatorname{MLP}([x_i;\, t_{\text{anchor}}])\bigr),
\end{equation}
where \(\sigma_i \in [0,1]\) determines how strongly each token relies
on the semantic anchor.  
This residual fusion allows the model to adaptively blend visual cues
with a domain-agnostic textual reference, producing more stable
cross-modal alignment and stronger semantic consistency between
3-D visual features and clinical language representations.

\begin{figure}[htpb]
  \centering
  \includegraphics[width=0.9\linewidth]{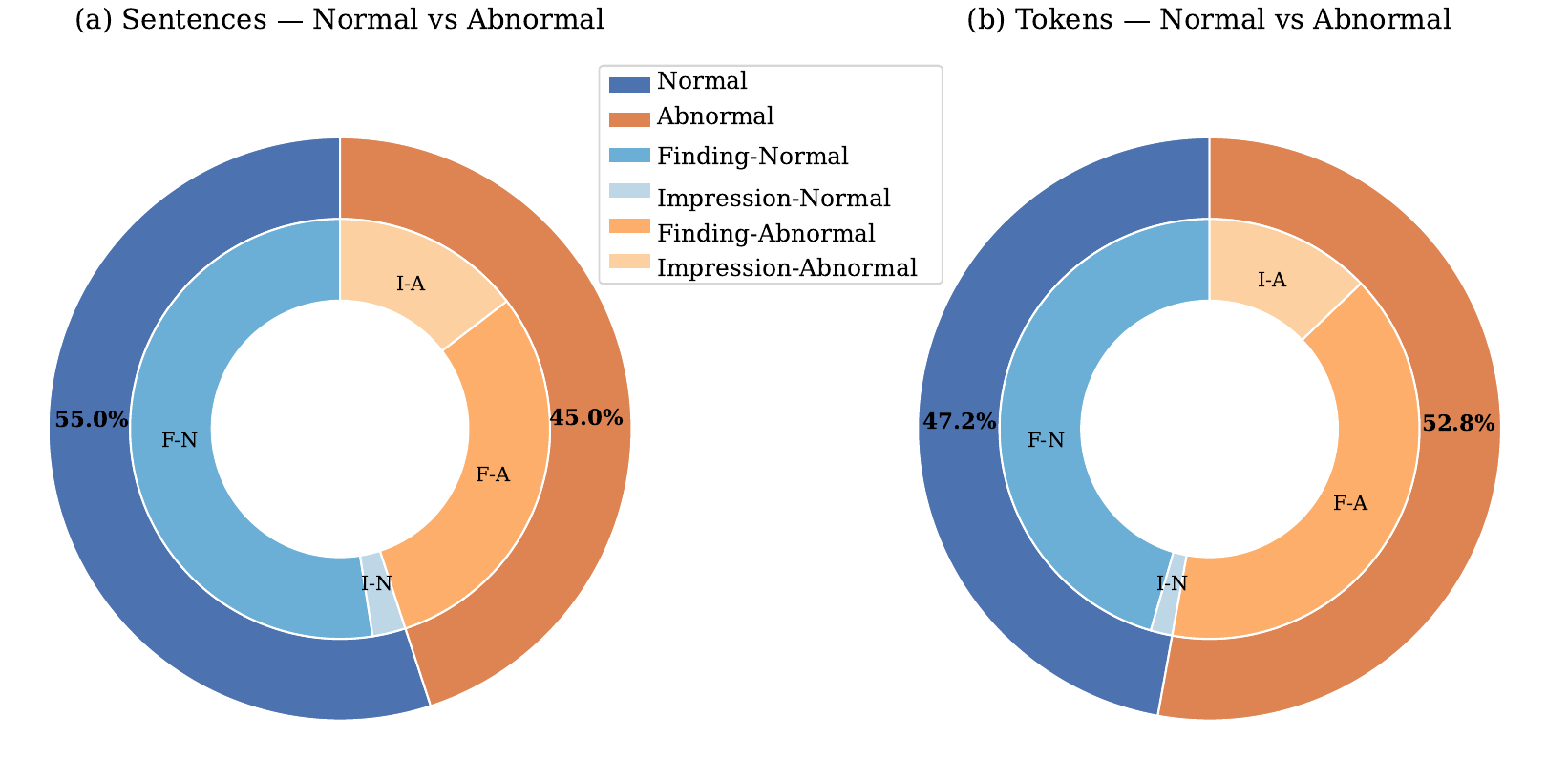} 
\caption{
Distribution of sentence- and token-level labels in CT-RATE reports.
The outer ring of each donut illustrates the overall proportion of \textit{normal} and \textit{abnormal} sentences (left) and tokens (right), showing that the two categories occur in comparable quantities.
The inner ring further decomposes each category into \textit{finding} and \textit{impression} components.
}
    \label{fig:data_stats}
\end{figure}

\subsection{Abnormality Re-Weighting}

During SFT, the model learns to align volumetric scans with clinical reports by predicting the next token in a sequence. In structured radiology reports, each anatomical region is described in a fixed order, regardless of whether a finding is present. Consequently, normal descriptions often prevail early on, whereas abnormal findings—though crucial—frequently appear later in the text. Our sentence-level analysis reveals that normal statements slightly outnumber abnormal ones (with a ratio of approximately 1:1.22 in CT-RATE dataset as shown in Figure~\ref{fig:data_stats}), a disparity that can be amplified by autoregressive language modeling. As a result, the model tends to reproduce normal patterns while overlooking clinically important abnormalities.

To mitigate this, we introduce an ARW strategy. First, each report is segmented into sentences; these are then labeled as either \emph{normal} or \emph{abnormal} using Deepseek-V3~\cite{guo2025deepseek}. During SFT, we adjust the loss for tokens belonging to abnormal sentences by a factor \(\lambda > 1\):

\begin{equation}
\mathcal{L} = - \sum_{t=1}^{T} \alpha_t \,\log P(\hat{y}_t = y_t),
\end{equation}

\begin{equation}
\alpha_t =
\begin{cases}
\lambda, & \text{if $y_t$ is from an abnormal sentence},\\
1,       & \text{otherwise},
\end{cases}
\end{equation}
where \(\alpha_t\) is the token-level weight. This re-weighting scheme aims to increase the model's sensitivity to clinically meaningful abnormalities, counteracting the tendency to overfit to repetitive, normal sentence patterns commonly seen at the beginning of reports.

\subsection{Rewarding Reasoning Consistency}

Although most RL-based methods for medical VLMs~\cite{lai2025med,pan2025medvlm} focus on generating structured responses and achieving high final-answer accuracy, they rarely verify whether the model’s reasoning process is grounded in genuine clinical evidence. To bridge this gap, we introduce a \emph{consistency reward} that supervises the semantic alignment between the model’s reasoning trace and the corresponding radiology report.

\paragraph{Task Formulation.} We formulate the abnormality diagnosis task as a Medical Multiple-choice VQA (MMVQA). Given a 3D CT scan and a clinical prompt, the model must select one correct answer from four options (A/B/C/D). The output is split into two parts: a step-by-step rationale enclosed in \texttt{<think>...</think>} tags, followed by a final decision in \texttt{<answer>...</answer>} tags (see Supplementary Material for details). This explicit separation of reasoning from the final conclusion facilitates the assessment and improvement of interpretability.

\paragraph{Consistency Reward.} Following SFT, we add a RL stage to optimize both accuracy and interpretability. While standard supervised learning encourages fidelity to reference answers, it does not directly promote high-quality diagnostic reasoning. To remedy this, we define a consistency reward that measures how well the \texttt{<think>} rationale aligns with the radiology report. A pre-trained reward model (RM-Mistral-7B~\cite{xiong2024iterative} quantifies semantic overlap; however, because this reward model is trained on general-domain data, its direct scores may reflect preferences that do not fully align with the medical domain. To reduce the impact of such biases while still leveraging the model’s linguistic understanding, we discretize its output into a small set of levels (e.g., 0, 0.5, 0.8, 1.0). This discretization mitigates domain mismatch and ensures that the reward signal remains robust and clinically relevant. Ablation study for discretization is provided in the Supplementary Material.

\paragraph{Overall Objective.}
The RL stage optimizes a sum of three rewards:
(i) a \emph{format reward} \(r_{\text{fmt}}\!\in\!\{0,1\}\) that checks compliance with the tags template,
(ii) an \emph{accuracy reward} \(r_{\text{acc}}\!\in\!\{0,1\}\) that verifies the chosen option, and
(iii) the \emph{consistency reward} \(r_{\text{con}}\) defined above.
The total reward is
\(
r = r_{\text{con}} +  r_{\text{fmt}} + r_{\text{acc}},
\)
which we maximize with GRPO~\cite{shao2024deepseekmath}: multiple responses \(\{o_k\}\sim\pi_\theta\) are sampled, their group-normalized advantages are computed, and a KL penalty keeps \(\pi_\theta\) close to a frozen reference policy \(\pi_{\text{ref}}\).

By integrating the consistency reward, our model not only selects the correct answer but also generates clinically grounded, logically coherent rationales, thereby yielding more trustworthy and interpretable diagnoses.
\section{Experiment}
\label{sec:experiment}

\subsection{Dataset}

\paragraph{CT-RATE~\cite{hamamci2024developing}}  
The publicly released CT\textsc{-}RATE corpus contains 50,188 non-contrast chest CT volumes (25,692 scans reconstructed twice with different kernels) from 21,304 patients, each paired with a radiology report. It provides 18 automatically extracted abnormality labels. We follow the \emph{official} split: 47,149 volumes for training and 3,039 for internal validation. During SFT, we jointly train on the report-generation task and 713,150 VQA samples generated by GPT-4~\cite{achiam2023gpt}. In the RL stage, diagnosis is framed as MMVQA: each query offers one true abnormality plus three random distractors. A subset covering 5 \% of patients (1,065 cases) is used for RL, while the official validation split evaluates both report generation and diagnostic reasoning.

\paragraph{RAD-ChestCT~\cite{draelos2021machine}} The RAD-ChestCT contains 3,630 publicly released chest CT volumes, each annotated with 84 abnormality labels. As the dataset lack radiology reports, it is used solely for evaluating diagnostic reasoning based on visual inputs. To ensure label consistency with CT-RATE, 27 abnormalities from RAD-ChestCT are mapped to the 18 abnormalities in CT-RATE, following the mapping protocol proposed in~\cite{lai2025bridged}.

\begin{table*}[h]
  \centering
  \setlength{\tabcolsep}{4mm}
  \renewcommand{\arraystretch}{1}
  \begin{tabular}{@{}c c c cccc@{}}
    \toprule
    Input Type & Method & LLM Backbone &
    \multicolumn{2}{c}{CT-RATE} & \multicolumn{2}{c}{RAD-ChestCT} \\
    \cmidrule(lr){4-5} \cmidrule(lr){6-7}
    & & & w think & w/o think & w think & w/o think \\
    \midrule

    \multirow{8}{*}{\centering {Montage}}
      & Qwen2.5-VL-3B~\cite{Qwen2.5-VL} & Qwen-2.5-3B 
      & 29.50 & 27.35 & 28.48 & 26.89 \\
      & Qwen2.5-VL-7B~\cite{Qwen2.5-VL} & Qwen-2.5-7B 
      & 29.23 & 28.98 & 26.35 & 26.54 \\
      & Qwen3-VL-4B-Instruct~\cite{yang2025qwen3} & Qwen-3-4B 
      & 29.93  & 33.52  & 29.42  &  34.48  \\
      & Qwen3-VL-4B-Thinking~\cite{yang2025qwen3} & Qwen-3-4B 
      &  28.00 &  32.77 &  30.85 &  34.88 \\

      & MedGemma-4B~\cite{sellergren2025medgemma} & Gemma-3-4B 
      & 24.00 & 24.06 & 27.75 & 28.61 \\
      & MedGemma-1.5-4B~\cite{sellergren2025medgemma} & Gemma-3-4B 
      & 23.07 & 21.71 & 25.60 & 24.40 \\
      & Lingshu-7B~\cite{xu2025lingshu} & Qwen-2.5-7B 
      & 28.54 & \underline{31.88} & 29.97 & \underline{35.79} \\
      & Med-R1~\cite{lai2025med} & Qwen-2-2B 
      & 29.32 & 27.96 & \underline{31.51} & 29.99 \\
      & MedVLM-R1~\cite{pan2025medvlm} & Qwen-2-2B 
      & 26.58 & 26.13 & 25.47 & 26.11 \\
    \midrule

    \multirow{8}{*}{\centering {DRR}}
      & Qwen2.5-VL-3B~\cite{Qwen2.5-VL} & Qwen-2.5-3B 
      & 25.70 & 23.09 & 28.36 & 25.17 \\
      & Qwen2.5-VL-7B~\cite{Qwen2.5-VL} & Qwen-2.5-7B 
      & 25.01 & 25.85 & 22.28 & 23.71 \\
      & Qwen3-VL-4B-Instruct~\cite{yang2025qwen3} & Qwen-3-4B 
      & 25.85  &  27.86 &  23.46 &  25.16 \\
      & Qwen3-VL-4B-Thinking~\cite{yang2025qwen3} & Qwen-3-4B 
      &  23.91 & 26.54  &  23.98 &  25.63  \\

      & MedGemma-4B~\cite{sellergren2025medgemma} & Gemma-3-4B 
      & 26.76 & 27.10 & 30.90 & 28.85 \\
      & MedGemma-1.5-4B~\cite{sellergren2025medgemma} & Gemma-3-4B 
      & 25.53 & 23.35 & 24.40 & 24.47 \\
      
      & Lingshu-7B~\cite{xu2025lingshu} & Qwen-2.5-7B 
      & 26.32 & 28.14 & 26.46 & 28.48 \\
      & Med-R1~\cite{lai2025med} & Qwen-2-2B 
      & 26.86 & 27.36 & 29.24 & 28.17 \\
      & MedVLM-R1~\cite{pan2025medvlm} & Qwen-2-2B 
      & 24.19 & 22.56 & 22.21 & 22.20 \\
    \midrule

    \multirow{6}{*}{\centering {Video}}
      & Qwen2.5-VL-3B~\cite{Qwen2.5-VL} & Qwen-2.5-3B 
      & 29.30 & 27.36 & 29.34 & 27.12 \\
      & Qwen2.5-VL-7B~\cite{Qwen2.5-VL} & Qwen-2.5-7B 
      & 26.89 & 26.63 & 25.32 & 24.50 \\
      & Qwen3-VL-4B-Instruct~\cite{yang2025qwen3} & Qwen-3-4B 
      &  26.34 &  28.65 &  28.74 &  30.30 \\
      & Qwen3-VL-4B-Thinking~\cite{yang2025qwen3} & Qwen-3-4B 
      &  26.45 & 28.95  & 27.06  & 29.07  \\

      & Lingshu-7B~\cite{xu2025lingshu} & Qwen-2.5-7B 
      & 26.17 & 29.14 & 29.25 & 35.19 \\
      & Qwen2.5-VL-3B-RL & Qwen-2.5-3B 
      & \underline{31.22} & 28.73 & 30.71 & 28.78 \\
    \midrule

    \multirow{7}{*}{\centering {Volume}}

      & RadFM~\cite{wu2025towards} & Llama-2-13B 
      & -- & 25.77 & -- & 25.00 \\
      & M3D~\cite{bai2024m3d} & Llama-2-7B 
      & -- & 22.68 & -- & 24.28 \\
      & E3D-GPT~\cite{lai2024e3d} & Vicuna-1.5-7B 
      & -- & 25.29 & -- & 25.15 \\
      & CT-CHAT~\cite{hamamci2024developing} & Llama-3.1-8B 
      & -- & 23.14 & -- & 20.52 \\
      & Med3DVLM~\cite{xin2025med3dvlm} & Qwen-2.5-7B 
      & -- & 21.59 & -- & 22.17 \\
      & \algname~(S1) & Qwen-2.5-3B 
      & 27.12 & 27.32 & 26.60 & 25.32 \\
      & \algname~(S2) & Qwen-2.5-3B 
      & \textbf{41.92} & \textbf{40.30} & \textbf{44.99} & \textbf{43.75} \\

    \bottomrule
  \end{tabular}
     \caption{Performance (\%) comparison on the MMVQA task using the CT-RATE and RAD-ChestCT validation datasets. ``w think'' and ``w/o think'' indicate whether reasoning instructions are applied during inference. ``S1'' refers to the stage 1 training with SFT, while ``S2'' denotes the stage 2 training with RL. A ``–'' indicates that the model is unable to produce valid multiple-choice answers under the ``w think'' setting due to its inability to follow reasoning-format prompts. Bold values represent the best performance, and underlined values indicate the second-best performance.}
  \label{tab:compare_other}
\end{table*}

\subsection{Implementation Details}

We first train a 3D CLIP~\cite{radford2021learning} model following BrgSA\cite{lai2025bridged}, a 3D extension of CLIP designed for aligning CT scans with their associated radiology reports. Specifically, we adopt a 3D ViT-base architecture~\cite{dosovitskiy2020image} as the vision encoder and train it on the CT-RATE dataset. After training, we retain only the pre-trained 3D ViT encoder, which is then used as the vision backbone in our subsequent 3D medical VLM. This vision encoder is integrated with the Qwen-2.5-3B\cite{yang2024qwen2} to enable downstream clinical reasoning tasks. All training and evaluation are conducted on a single H20 server with 8 NVIDIA GPUs.

In stage~1, We first perform SFT on caption and VQA data, updating only the projection layer. This approach enhances vision-text alignment while preserving the semantic capacity of the large language model.  $\lambda$ is empirically set to 1.10 in ARW(see Supplementary Material). In stage~2, We then introduce RL on the MMVQA dataset that jointly maximizes a sum of three rewards. Optimizing these rewards together boosts diagnostic correctness, template compliance, and clinically grounded reasoning in a single training pass.

We use accuracy to evaluate the MMVQA task. For report generation, we adopt BLEU, METEOR, ROUGE, and CIDEr~\cite{lin2014microsoft}. In addition, we include the semantic metrics BERT F1 score (BERT\_F1)~\cite{zhang2019bertscore} and LLM score (LLM\_Score) using DeepSeek-V3~\cite{guo2025deepseek}.


\subsection{Comparison with State-of-the-art Methods}

Current VLMs are broadly categorized into three types based on the input data: (1) 2D image input, (2) video input, and (3) 3D volumetric data input. In this study, we design three input processing strategies for 3D CT scans, based on the types of data supported by existing VLMs:

First, for \textbf{3D-to-2D Conversion}, we design two methods: one converts the 3D CT scan data into a \textbf{multi-slice montage}, where multiple slices are stitched into a single 2D image, preserving detailed anatomical structures; the second uses Digital Reconstructed Radiographs (\textbf{DRR}) to project the 3D volumetric data along three orthogonal directions, simulating X-ray images from three views (see Supplementary Material for details). For \textbf{Video Input}, we treat the depth dimension as the time axis, organizing CT slices along the depth dimension into video frames, simulating the dynamic changes of slices as the depth increases. Lastly, \textbf{3D Volume Input} retains the full 3D structural information, including height, width, and depth, ensuring complete anatomical details are presented.

As shown in Table~\ref{tab:compare_other}, we evaluate models under these different input types. The results indicate that for the 2D input scenario, models using the montage strategy perform better, except for MedGemma. This is mainly because DRR loses more information due to the simplification process, while the montage strategy retains more slice details. MedGemma, however, performs well with DRR, likely due to its training on a large amount of X-ray data, which gives it an advantage on DRR inputs. In the case of video input, although this method simulates the dynamic relationships between slices, it does not bring a significant performance improvement compared to the montage strategy. This is likely due to information loss from frame sampling in video inputs, which limits the model’s ability to capture fine-grained details. This suggests that for 3D medical imaging, the montage strategy is more effective than both DRR and video input, especially in capturing spatial details.

Additionally, we apply GRPO training to the Qwen2.5-VL-3B on video input, named Qwen2.5-VL-3B-RL, and compare it with the Qwen2.5-VL-3B model. The results show a slight performance improvement after applying RL, with CT-RATE increasing from 29.30 to 31.22 and RAD-ChestCT from 29.34 to 30.71. However, compared to our \algname~(S1) model, which already contains a solid foundation of medical knowledge after SFT, the performance improvement is much more pronounced. \algname's performance increases substantially, with CT-RATE rising from 27.12 to 41.92 and RAD-ChestCT from 26.60 to 44.99. This result further validates that models with a medical knowledge base can better leverage RL to enhance reasoning abilities, thus achieving significant performance gains.

Our experiments also reveal an important issue with existing 3D medical VLMs on 3d volume: overfitting, which leads to a decline in instruction-following ability. Many existing methods experience overfitting during LLM training, which results in poor performance when handling out-of-distribution instructions. The outputs fail to follow the given prompts accurately, leading to significantly lower performance (see Supplementary Material for details). To ensure a fair evaluation under these conditions, the results reported in our tables are post-processed by selecting the option with the highest cosine similarity to the model's raw output as the final answer. This issue highlights the limitations of current 3D medical VLMs in following multimodal instructions, and underscores the advantages of our approach in addressing this challenge.

\begin{table*}[htpb]
\centering
\setlength{\tabcolsep}{1.mm}
\begin{tabular}{@{}cccccccccccccc@{}}
\toprule
RAM & ARW 
    & B‑1 & B‑2 & B‑3 & B‑4 
    & METEOR 
    & R‑1 & R‑2 & R‑L 
    & CIDEr 
    & BERT\_F1 & LLM\_Score 
    & Mean \\
\midrule

\checkmark & \checkmark 
            & \textbf{40.93} & \textbf{30.98} & \textbf{25.19} & \textbf{21.55}
            & \textbf{36.18}
            & \textbf{56.63} & \textbf{34.50} & \textbf{39.84}
            & \textbf{23.86} 
            & \textbf{88.70} & \textbf{80.48}
            & \textbf{43.53} \\

\checkmark &  $\times$  & 38.51$^{*}$ & 28.96$^{*}$ & 23.40$^{*}$ & 19.92$^{*}$ & 35.16$^{*}$ & 55.94$^{*}$ & 33.59$^{*}$ & 39.11$^{*}$ & 20.22 & 88.47$^{*}$ & 79.50$^{\dagger}$ & 42.07 \\

$\times$  & \checkmark & 39.83$^{*}$ & 29.89$^{\dagger}$ & 24.08$^{\dagger}$ & 20.42$^{\dagger}$ & 35.49$^{*}$ & 56.19$^{*}$ & 33.59$^{\dagger}$ & 38.83$^{\dagger}$ & 22.51 & 88.41$^{\dagger}$ & 79.49$^{\dagger}$ & 42.61 \\

$\times$ & $\times$ & 37.77$^{\dagger}$ & 28.07$^{\dagger}$ & 22.46$^{\dagger}$ & 18.97$^{\dagger}$  & 34.43$^{\dagger}$  & 55.19$^{\dagger}$ & 32.63$^{\dagger}$ & 37.79$^{\dagger}$  & 22.07$^{*}$ & 88.21$^{\dagger}$ & 78.48$^{\dagger}$  & 41.46 \\

\bottomrule
\end{tabular}
\caption{Ablation study of the proposed modules on report generation (metrics in \%). Abbreviations: B-n = BLEU-n; R-n = ROUGE-n. $^{*}$ indicates $p < 0.05$, and $^{\dagger}$ indicates $p < 0.005$ vs.\ the full model with both modules, using a paired \textit{t}-test.}

\label{tab:ablation_final_percent_corrected}
\end{table*}

\begin{table}[htpb]
  \centering
  \setlength{\tabcolsep}{1.4mm}
  \begin{tabular}{@{}l l c c c c@{}}
    \toprule
    Bridge & LoRA
      & \multicolumn{2}{c}{CT‑RATE} 
      & \multicolumn{2}{c}{RAD‑ChestCT} \\
    \cmidrule(lr){3-4} \cmidrule(lr){5-6}
     & LLM & w think & w/o think & w think & w/o think \\
    
    \midrule
    \checkmark &  $\times$     & 27.12 & 27.32 & 26.60 & 25.32 \\
    \checkmark & \checkmark             & --     & \textbf{27.81} & --     & \textbf{25.46} \\
    \bottomrule
  \end{tabular}
  \caption{Performance (\%) of SFT on the MMVQA task under different training modules.}
\label{tab:bridge_only}
\end{table}

\subsection{Ablation Study}
\label{sec:as}

Table \ref{tab:ablation_final_percent_corrected} contrasts the full model with three ablated variants to diagnose the roles of the ARW and the RAM. The complete system achieves the highest score on every metric; once either component is removed, scores fall significantly ($p<0.05$ or $p<0.005$), confirming that the two modules provide complementary rather than redundant benefits.

When ARW is removed while RAM is retained, the largest absolute drop occurs in CIDEr (–3.64). Because CIDEr assigns higher weights to low-frequency, clinically specific n-grams, this sharp decrease indicates that many rare abnormal terms are now missing or incorrectly generated. The observation supports ARW’s intended effect: by up-weighting abnormal sentences during training, it helps the decoder preserve detailed pathology descriptions that would otherwise be under-represented.

Conversely, eliminating RAM but keeping ARW yields the steepest declines in the embedding-based measures BERT\_F1 (–0.29) and LLM\_Score (–0.99). These metrics evaluate semantic fidelity and factual consistency across the entire report, so their sensitivity to RAM’s absence suggests that RAM chiefly secures global vision-language alignment. Disabling both modules pushes all scores to their lowest levels, confirming that robust report generation hinges on the joint action of ARW for abnormal focus and RAM for holistic cross-modal grounding.

To verify that restricting SFT to the vision–language Bridge is sufficient, we conduct the experiment summarized in Table \ref{tab:bridge_only}. Fine-tuning the Bridge alone preserves the model’s reasoning capability, yielding 27.1\% and 26.6\% accuracy with reasoning on CT-RATE and RAD-ChestCT, while direct-answer scores remain comparable. When we also adapt LLM weights with LoRA, reasoning disappears and direct-answer accuracy rises by no more than 0.5 \%—an outcome that underscores how, under limited training data, altering the LLM’s parameters degrades instruction compliance more than it boosts performance. Hence, we confine SFT to Bridge-only tuning and leave full-model updates to the subsequent RL phase.

\subsection{Effectiveness of Consistency Reward}

Table~\ref{tab:reward_radiologist} summarizes the evaluation results from two complementary judges—a board-certified radiologist with ten years of clinical experience and the large-language model Deepseek-V3—who independently assessed the model’s reasoning quality. Both judges scored four dimensions of reasoning: Clinical Correctness, Clarity, Differential Diagnosis, and Conciseness.

Incorporating the consistency reward improves model performance across every dimension for both judges. For the radiologist, scores increase by {+0.21 to +0.40}, with the largest gain in \emph{Clinical Correctness} (+0.40). Deepseek-V3 reports even larger improvements of {+0.21 to +0.90}, with the largest gain in \emph{Differential Diagnosis} (+0.90).

All gains are statistically significant ($p<0.0001$), demonstrating that the consistency reward yields reasoning that is more accurate, clearer, and more clinically grounded. The full evaluation prompt used for Deepseek-V3 is provided in the Supplementary Material.

\begin{table}[htpb]
\centering
\setlength{\tabcolsep}{0.8mm}
\begin{tabular}{@{}l|c|cccc@{}}
\toprule
Judge &  Consistency & Clinical & Clarity & Diff. Dx & Concis \\
\midrule
Radiologist  &  $\times$   & 3.13 & 3.14 & 3.08 & 3.26  \\
Radiologist & \checkmark & \textbf{3.53}$^{*}$ & \textbf{3.48}$^{*}$ & \textbf{3.35}$^{*}$ & \textbf{3.50}$^{*}$ \\
 \midrule
Deepseek  & $\times$ &   3.53 & 3.53 & 2.64 & 3.69   \\
Deepseek  & \checkmark & \textbf{4.29}$^{*}$ & \textbf{4.08}$^{*}$ & \textbf{3.54}$^{*}$ & \textbf{3.90}$^{*}$   \\
\bottomrule
\end{tabular}
\caption{Evaluation scores across different reward strategies. 
Abbreviations: Clinical (Clinical Correctness), Clarity (Reasoning Clarity), Diff.\,Dx (Differential Diagnosis), Concis (Conciseness).
All metrics range from 0 to 5 and are assessed independently by a board-certified radiologist and Deepseek-V3. 
$^{*}$ indicates $p < 0.0001$ vs.\ the baseline.}
\label{tab:reward_radiologist}
\end{table}

\section{Limitation}
\label{sec:limitation}

Our study has two main limitations. First, the training data are limited in scale, \emph{anatomical coverage}, and diversity: we rely primarily on public chest CT--report pairs. While synthetic VQA increases \emph{textual} variety, it does not enlarge the \emph{image} pool, broaden organ/region coverage, or capture multi-center heterogeneity, which may constrain robustness and out-of-distribution generalization. In future work, we will curate larger, multi-center CT--report cohorts, quantify data-scaling effects, and systematically expand to additional organs and imaging modalities by collecting substantially more images (and reports) beyond the thorax.

Second, the supervision for clinical reasoning remains insufficient. The model’s reasoning traces largely reflect intrinsic language ability rather than clinician-authored, auditable evidence chains; the reward model is predominantly general-domain, risking fluent yet weakly evidenced explanations. Going forward, we will collaborate with radiologists to construct high-quality, rubric-based rationales and use them to fine-tune the reward model and to constrain SFT+RL training, thereby improving the clinical faithfulness and interpretability of reasoning outputs.

\section{Conclusion}
\label{sec:conclusion}

We introduce \algname, the first 3D vision–language system to use RL for stronger clinical reasoning. RAM and ARW mitigate volumetric-data biases during SFT, while a consistency reward in RL further enforces coherent, clinically sound diagnoses. On CT-RATE and RAD-ChestCT, \algname sets a new state-of-the-art and yields more logical, interpretable reports, demonstrating a practical path toward transparent, trustworthy AI in 3D medical imaging.

{
    \small
    \bibliographystyle{ieeenat_fullname}
    \bibliography{main}
}



\end{document}